\documentclass[11pt,a4paper]{article}

\usepackage[top=0.9in,bottom=0.9in,left=0.9in,right=0.9in]{geometry}
\usepackage{times}
\usepackage{soul}
\usepackage[utf8]{inputenc}
\usepackage[small]{caption}
\usepackage{amsmath,amssymb}
\usepackage{bm}
\newcommand{\R}{\mathbb{R}}
\newcommand{\T}{\mathsf{T}}

\usepackage{algorithm}
\usepackage{algorithmic}

\usepackage{url}
\usepackage{caption}
\usepackage{here}
\usepackage{comment}
\usepackage{afterpage,array,rotating}
\usepackage{multicol}
\newcolumntype{Y}{>{\centering\arraybackslash}p{80pt}}

\title{Spectral Feature Scaling Method \\for Supervised Dimensionality Reduction}

\author{Momo Matsuda$^1$, Keiichi Morikuni$^1$, Tetsuya Sakurai$^{1,2}$
  \\ 
  $^1$ University of Tsukuba \\
  $^2$ JST/CREST\\
  matsuda@mma.cs.tsukuba.ac.jp
}
\date{}

\begin{document}

\maketitle

\begin{abstract}
  Spectral dimensionality reduction methods enable linear separations of complex data with high-dimensional features in a reduced space. However, these methods do not always give the desired results due to irregularities or uncertainties of the data. Thus, we consider aggressively modifying the scales of the features to obtain the desired classification. Using prior knowledge on the labels of partial samples to specify the Fiedler vector, we formulate an eigenvalue problem of a linear matrix pencil whose eigenvector has the feature scaling factors. The resulting factors can modify the features of entire samples to form clusters in the reduced space, according to the known labels. In this study, we propose new dimensionality reduction methods supervised using the feature scaling associated with the spectral clustering. Numerical experiments show that the proposed methods outperform well-established supervised methods for toy problems with more samples than features, and are more robust regarding clustering than existing methods. Also, the proposed methods outperform existing methods regarding classification for real-world problems with more features than samples of gene expression profiles of cancer diseases. Furthermore, the feature scaling tends to improve the clustering and classification accuracies of existing unsupervised methods, as the proportion of training data increases.
\end{abstract}

\section{Introduction}
Consider clustering a set of data samples with high-dimensional features into mutually exclusive subsets, called clusters, and classifying these clusters when given prior knowledge on the class labels of partial samples. These kinds of problems arise in pathological diagnoses using gene expression data \cite{gene}, the analysis of chemical sensor data \cite{sensor}, the community detection in social networks \cite{network}, the analyses of neural spike sorting \cite{spike}, and so on \cite{detection}. Spectral clustering is an unsupervised technique that projects the data samples to the eigenspace of the Laplacian matrix, in which the data samples are linearly separated if they are well clustered. This technique is more effective in complicated clustering than existing methods. For example, the $k$-means algorithm induces partitions using hyperplanes, but may fail to give satisfactory results due to irregularities and uncertainties of the features.

Here, we consider aggressively modifying the scales of the features to improve the separation in the reduced space, and obtain the desired classification. We derive the factors for scaling the features (scaling factors) using a learning machine based on spectral dimensionality reduction; namely, the inputs are unscaled and labeled samples, and the outputs are the scaling factors. To this end, by exploiting the prior knowledge on the labels of partial samples, we specify the Fiedler vector and reformulate the Laplacian eigenproblem as an eigenproblem of a linear matrix pencil whose eigenvector has the scaling factors. The obtained factors can modify the features of the entire samples to form clusters in the reduced dimensionality space according to the known labels. Thus, we use the prior knowledge to implicitly specify the phenomenon of interest, and supervise dimensionality reduction methods for the scaling factors. These approaches yield the desired clusters, incorporated with unsupervised spectral clustering, for test data scaled by the obtained factors for training data. Numerical experiments on artificial data and real-world data from gene expression profiles show that the feature scaling improves the accuracy of the spectral clustering. In addition, the spectral dimensionality reduction methods supervised using the feature scaling outperform existing methods in some cases, and are more robust than existing methods.

We review related work on supervised spectral dimensionality reduction methods. Learning a similarity matrix from training data is effective in spectral clustering \cite{super1,super2}. The proposed classification method can be considered a supervised method, incorporated with the kernel version of the locality preserving projections (LPP) \cite{lpp}. LPP is comparable with the linear discriminant analyisis (LDA) \cite{lda}, the local Fisher discriminant analysis (LFDA) \cite{lfda}, and the locality adaptive discriminant analysis (LADA) \cite{lada}. Kernel versions of LDA, LPP and LFDA aim at nonlinear dimensionality reduction. These methods have semi-supervised variants \cite{uni,semisuper,Cui}.

\subsection{Spectral Clustering}

The underlying idea of the proposed methods follows from spectral clustering, which is induced by graph partitioning, where the graph is weighted and undirected. The weight $w_{i,j}$ of the edge between nodes $i$ and $j$ represents the similarity between samples $i$ and $j$ for $i, j = 1, 2, \ldots, n$, where $n$ is the number of samples. By dividing the graph into mutually exclusive subgraphs, the corresponding samples form clusters. This discrete problem can be continuously relaxed to a matrix eigenvalue problem. Let
\begin{align*}
  W&=\lbrace w_{i,j}\rbrace\in\R^{n\times n}, \\
  D&=\mathrm{diag}\left(d_1,d_2,\ldots,d_n\right)\in\R^{n\times n} \quad \mbox{ with } \quad d_i = \sum_{i=1}^n w_{i,j},  
\end{align*}
$\bm{e}\in\R^n$ be a vector with all elements $1$, and $\bm{t}$ be an indicator vector with the $i$th entry $t_i = 1$ if sample $i$ is in a cluster, and $t_i = -1$ if sample $i$ is in another cluster. Then, the constrained minimization of the Ncut function \cite{sc}
\begin{equation}
  \min_{\bm{v}} \frac{\bm{v}^{\mathsf{T}} (D - W) \bm{v}}{\bm{v}^{\mathsf{T}}D\bm{v}}, \quad \mbox{subject to} \quad \bm{e}^{\mathsf{T}}D\bm{v} = 0
  \label{eq:j}
\end{equation}
over $v_i \in \{ 1, -b \}$ with $b = \sum_{t_i > 0}d_i \slash \sum_{t_i<0}d_i$, is relaxed to finding the Fiedler vector $\bm{v} \in \R^n \setminus \{ \bm{0} \}$ \cite{fiedler} associated with the smallest eigenvalue of the constrained generalized eigenvalue problem
\begin{equation}
  L\bm{v} = \lambda D \bm{v}, \quad \mbox{subject to} \quad \bm{e}^{\mathsf{T}}D\bm{v} = 0,
  \label{eq:lap}
\end{equation}
where $L = D - W$ is a Laplacian matrix and $\lambda \in \R$. The latter problem (\ref{eq:lap}) is tractable, while the former (\ref{eq:j}) is NP-hard. Moreover, a few eigenvectors of (\ref{eq:lap}) form clusters that are separable using hyperplanes if the samples form well-separated clusters. A sophisticated algorithm for (\ref{eq:lap}) is nearly linear time \cite{time}.

\section{Proposed Method}
Irregular scales and uncertainty of features prevent well-established dimensionality reduction methods from clustering data samples into the desired disjoint subsets. In spectral clustering, the Ncut function is popular by virtue of the nonlinear separability, but is not versatile. To cope with these issues, we propose a remedy to aggressively modify the scales of the features in the eigenspace where a linear separation works, based on prior knowledge of the partial $n$ samples $X = \left[\bm{x}_1, \bm{x}_2, \ldots, \bm{x}_n\right]^\T, \bm{x}_i\in\R^m$ with $m$ features. If the classes of partial samples are known, we can estimate the entries of the Fiedler vector $\bm{v}\in\R^n\backslash\lbrace\bm{0}\rbrace$ of the Laplacian eigenvalue problem
\begin{equation}
  L_s \bm{v} = \lambda_s D_s \bm{v}, \quad \bm{e}^\T D_s \bm{v} = \bm{0}, \quad \lambda_s \in \R,
  \label{eq:LS}
\end{equation}
where $L_s = D_s - W_s$, and $W_s = \lbrace w_{i,j}^{(s)} \rbrace \in \R^{n\times n}$ depends on the feature scaling factors $\bm{s}\in\R^m$. We obtain the scaling factors $\bm{s}\in\R^m$ in (\ref{eq:LS}) by solving an eigenproblem of a linear matrix pencil. Then, applying the obtained scaling factors to the features of the entire data $Y = \left[X^{\T}, R^\T\right]^\T\in\R^{N \times m}$ with $N$ samples, we can extend the prior knowledge on the partial samples to overall samples, where $R\in\R^{(N-n)\times m}$ denotes the remaining data. In statistical terms, the scaling changes the mean $\bar{y}_j = \frac{1}{n}\sum_{i=1}^n y_{i,j}$ and the variance $\sigma_j^2 = \frac{1}{n} \sum_{i=1}^n (y_{i,j} - \bar{y}_j)^2$ of the $j$th feature to $s_j^{1/2}\bar{y}_j$ and $|s_j|\sigma_j^2$, respectively, for $j = 1, 2, \ldots, m$. Note that we allow the scaling factors to have negative values. See \cite{negative}. Finding appropriate scaling factors is also considered metric learning \cite{metric}. Recall that the center of a cluster associated with a Bregman divergence is equal to the centroid of the samples of the cluster \cite{diver}.

Now, we reformulate (\ref{eq:LS}) as another eigenproblem to extract the scaling factors $s_i\in\R,i = 1, 2, \ldots, m$ as an eigenvector. Denote the scaling matrix by $S^{\frac{1}{2}}=\mathrm{diag}\left(s_1, s_2, \ldots, s_m\right)^{\frac{1}{2}}\in~\R^{m \times m}$ and denote the $(i,j)$ entry of the similarity matrix $W_s$ for the scaled data $XS^{1/2}$ by
\begin{eqnarray}
  w_{i,j}^{(s)} = \begin{cases}
    1 - \frac{\left( \bm{x}_i - \bm{x}_j \right)^{\T} S \left( \bm{x}_i - \bm{x}_j \right)}{2\sigma^2}  = 1 - \frac{\bm{s}^{\mathsf{T}} \bm{x}_{i,j}}{2\sigma^2}, \\
    \quad \simeq \exp\left(- \frac{\left( \bm{x}_i - \bm{x}_j \right)^{\T} S \left( \bm{x}_i - \bm{x}_j \right)}{2\sigma^2}\right),& i \neq j,\\
    & \\
    0, & i = j,
  \end{cases}
         \label{eq:weight}
\end{eqnarray}
where $\bm{s} = \left[ s_1, s_2, \ldots, s_m\right]^\T\in\R^m$ and the $k$th entry of $\bm{x}_{i,j}\in\R^m$ is $\left(x_{i,k} - x_{j,k}\right)^2$. Here, we used the first-order approximation of the exponential function $\exp\left(-x\right) \approx 1 - x \quad\mbox{for}\quad0<x<1.$
Then, the $i$th row of $W_s$ is
\begin{equation*}
  \begin{split}
    \bm{w}_i^{(s)\T}& = \left[1,\ldots,1,0,1,\ldots,1\right]-\frac{\bm{s}^\T}{2\sigma^2}\left[\bm{x}_{i,1},\bm{x}_{i,2},\ldots,\bm{x}_{i,n}\right]\\
    & = \tilde{\bm{e}}_i^\T-\bm{s}^\T X_i,
  \end{split}
\end{equation*}
where $\tilde{\bm{e}}_i$ is the $n$-dimensional vector with the $i$th entry equal to zero and the remaining entries ones, and $X_i = \frac{1}{2\sigma^2}\left[\bm{x}_{i,1},\bm{x}_{i,2},\ldots,\bm{x}_{i,n}\right] \in\R^{m\times n}.$
Hence, we have
\begin{equation*}
  W_s\bm{v} = \left[\tilde{\bm{e}}_1^\T, \tilde{\bm{e}}_2^\T, \ldots, \tilde{\bm{e}}_n^\T \right]^\T \bm{v} - \left[X_1\bm{v}, X_2\bm{v}, \ldots, X_n\bm{v}\right]^\T \bm{s}.
\end{equation*}
Let $\hat{\bm{x}}_i = \frac{1}{2\sigma^2}\left(\bm{x}_{i,1} + \bm{x}_{i,2} + \cdots + \bm{x}_{i,n}\right).$
Then, the $i$th diagonal entry $d_i^{(s)}$ of $D_s$ is
\begin{equation*}
  d_i^{(s)} = \sum_{j=1}^nw_{i,j}^{(s)} = (n-1) - \bm{s}^\T\hat{\bm{x}}_i.
\end{equation*}
Hence, denoting the Fiedler vector by $\bm{v} = \left[ v_1, v_2, \ldots, v_n \right]^\T$, we have
\begin{eqnarray*}
  D_s\bm{v} = (n-1)\bm{v} - \left[v_1\hat{\bm{x}}_1, v_2\hat{\bm{x}}_2, \ldots, v_n\hat{\bm{x}}_n\right]^\T \bm{s}.
\end{eqnarray*}
Thus, (\ref{eq:LS}) is written as
\begin{eqnarray*} 
  L_s\bm{v} = \lambda_sD_s\bm{v} &\Leftrightarrow& W_s\bm{v} = (1 - \lambda_s)D_s\bm{v}\\
                                 &\Leftrightarrow&
                                                   \begin{bmatrix}
                                                     A & \bm{\alpha}\\
                                                   \end{bmatrix}
  \begin{bmatrix}
    \bm{s}\\
    -1\\    
  \end{bmatrix} = \mu
  \begin{bmatrix}
    B & \bm{\beta}
  \end{bmatrix}
        \begin{bmatrix}
          \bm{s}\\
          -1
        \end{bmatrix},
\end{eqnarray*}
where $\mu = 1 - \lambda_s$, 
\begin{eqnarray}  A =
  \begin{bmatrix}
    \left(X_1\bm{v}\right)^\T\\
    \left(X_2\bm{v}\right)^\T\\
    \vdots\\
    \left(X_n\bm{v}\right)^\T\\
  \end{bmatrix}, \quad B =
  \begin{bmatrix}
    v_1\hat{\bm{x}}_1^\T\\
    v_2\hat{\bm{x}}_2^\T\\
    \vdots\\
    v_n\hat{\bm{x}}_n^\T\\
  \end{bmatrix}\in\R^{n\times m},
  \label{eq:AB}
\end{eqnarray}
\begin{eqnarray}
  \bm{\alpha} = \left[\tilde{\bm{e}}_1, \tilde{\bm{e}}_2, \ldots, \tilde{\bm{e}}_n\right]^\T\bm{v}\in\R^n,\quad \bm{\beta} = (n-1)\bm{v}\in\R^n.
  \label{eq:so}
\end{eqnarray}
Furthermore, the feature scaling factors $\bm{s}$ have the constraint
\begin{eqnarray*}
  &&\bm{e}^\T D_s\bm{v} = \sum_{i=1}^n\left((n-1)v_i - v_i\bm{s}^\T \hat{\bm{x}}_i\right) = 0\\
  &\Leftrightarrow& \left(\sum_{i=1}^nv_i\hat{\bm{x}}_i^\T\right)\bm{s} - (n-1)\sum_{i=1}^n v_i = 0.
\end{eqnarray*}
Therefore, we can obtain the scaling factors $\bm{s}$ by solving the generalized eigenproblem of a linear matrix pencil
\begin{eqnarray}
  \begin{bmatrix}
    A & \bm{\alpha}\\
    \bm{\gamma}^\T & \rho
  \end{bmatrix}
                     \begin{bmatrix}
                       \bm{s}\\
                       -1
                     \end{bmatrix} = \mu
  \begin{bmatrix}
    B & \bm{\beta}\\
    \bm{0}^\T & 0    
  \end{bmatrix}
                \begin{bmatrix}
                  \bm{s}\\
                  -1
                \end{bmatrix}, \quad \bm{s}\in\R^m, \quad \mu\in\R,
  \label{eq:rec}
\end{eqnarray}
where
\begin{eqnarray}
  \bm{\gamma} = \sum_{i=1}^nv_i\hat{\bm{x}}_i\in\R^m, \quad \rho = (n-1)\sum_{i=1}^nv_i\in\R.
  \label{eq:gammarho}
\end{eqnarray}
It follows from the equality $\left( A - B \right)^\T \bm{e} = \bm{0}$ that there exists a real number $\mu$ that satisfies (\ref{eq:rec}). To solve (\ref{eq:rec}), we can use the contour integral method \cite{ss} for $n < m$; otherwise, we can also use the minimal perturbation approach \cite{svd}. When $\bm{s}$ is obtained as a complex vector, its real and imaginary parts are also eigenvectors of (\ref{eq:rec}). We took the real part in the experiments given in section~\ref{expe}. For $\mu = 1 - \lambda$, we adopt the eigenvector of (\ref{eq:rec}) associated with the eigenvalue $\mu$ closest to one as the scaling factors. Then, we apply the squared feature scaling factors $\bm{s}^{1/2}$ to the entire data $Y$, such that $Z = YS^{1/2}$. Finally, we perform the spectral clustering or classification on the scaled data $Z \in\R^{N\times m}$ by solving the Laplacian eigenvalue problem
\begin{equation*}
  L' \bm{u} = \lambda' D' \bm{u}, \quad \bm{e}^{\T}\!D'\bm{u} = 0, \quad \bm{u}\in\mathbb{R}^N\setminus\lbrace\bm{0}\rbrace, \quad \lambda' \in \mathbb{R} 
\end{equation*}
for a few eigenvectors corresponding to the smallest eigenvalues, where $L'=D'-W'\in\R^{N \times N}, W' = \lbrace w'_{i,j} \rbrace \in \R^{N \times N}$,
\begin{equation}
  w'_{i,j} = \exp\left(-\frac{{\| \bm{z}_i - \bm{z}_j \|_2}^2 }{2 \sigma^2}\right),\quad  i, j = 1, 2 \ldots, N,
  \label{eq:w}
\end{equation}
$\bm{z}_i \in \R^m$ is the $i$th row of $Z$, $d'_i=\sum_{j=1}^n w'_{i,j}$, and $D'=\mathrm{diag} \left(d'_1, d'_2, \ldots, d'_N\right)$. We summarize the procedures of the proposed methods in Algorithm~\ref{alg1}.

\begin{algorithm}[!ht]             
  \caption{Spectral clustering/classification methods supervised using the feature scaling}        
  \label{alg1}                          
  \begin{algorithmic}[1]
    \REQUIRE Training data $X \in \mathbb{R}^{n \times m}$, entire data $Y \in \mathbb{R}^{N \times m}$, estimated Fiedler vector $\bm{v}$, dimension of the reduced space $\ell$.
    \STATE Compute $A, B, \bm{\alpha}, \bm{\beta}, \bm{\gamma}, \rho$ in (\ref{eq:AB}), (\ref{eq:so}), (\ref{eq:gammarho}).
    \STATE Solve (\ref{eq:rec}) for an eigenvector $\bm{s} = [s_1, s_2, \ldots, s_m]^{\mathsf{T}}$.
    \STATE Set the scaling matrix $S^{1/2} = \mathrm{diag}(s_1, s_2, \ldots, s_m)^{1/2}$.
    \STATE Compute the scaled data matrix $Z = YS^{1/2}$.
    \STATE Compute the similarity matrix $W' \in \mathbb{R}^{N \times N}$ (\ref{eq:w})  and $D'$, using the $k$-nearest neighbors, and set $L' = D' - W'$.
    \STATE Solve $L'\bm{u} = \lambda' D' \bm{u}$ for the eigenvectors $\bm{u}_1,\bm{u}_2 \ldots \bm{u}_\ell$ corresponding to the $\ell$ nonzero smallest eigenvalues.
    \STATE Cluster/classify the reduced data samples $\left[\bm{u}_1,\bm{u}_2,\ldots,\bm{u}_\ell\right]\in\R^{N \times\ell}$ rowwise by using the $k$-means algorithm/the one-nearest neighborhood method, respectively.
  \end{algorithmic}
\end{algorithm}

\section{Numerical experiments}\label{expe}
Numerical experiments on test problems were performed to compare the proposed methods with existing methods in terms of accuracy. For clustering problems, we compared the proposed method (spectral clustering method supervised using the feature scaling, SC-S) with the unsupervised spectral clustering method (SC) \cite{sc}, LPP, its kernel version (KLPP) \cite{lpp}, and two supervised methods, LFDA and its kernel version (KLFDA) \cite{lfda}. For classification problems, we compared the proposed method (spectral classification method supervised using the feature scaling, SC-S) with LPP, KLPP, LFDA, and KLFDA. All programs were coded and run in Matlab 2016b. We employed the LPP code in the Matlab Toolbox for Dimensionality Reduction\footnote{\url{https://lvdmaaten.github.io/drtoolbox/}} and the LFDA and KLFDA codes in the Sugiyama-Sato-Honda Lab site\footnote{\url{http://www.ms.k.u-tokyo.ac.jp/software.html}}. In the compared methods except for LFDA, we chose the optimal value of the parameter $\sigma$ among $1, 10^{\pm1}$, and $10^{\pm2}$ to achieve the best accuracy. We formed the similarity matrix in (\ref{eq:w}) by using the seven-nearest neighbor and taking the symmetric part. We set the dimensions of the reduced space to one for clustering problems, and one, two and three for classification problems. It is known that using one eigenvector is sufficient for binary clustering if the clusters are well-separated \cite{theroy}. The reduced data samples to a low-dimension were clustered using the $k$-means algorithm, and classified using the one-nearest neighborhood method.

The performance measures we used were the normalized mutual information (NMI) \cite{nmi} and the Rand index (RI) \cite{randindex} for evaluating the clustering methods, and RI for evaluating the classification methods. Let $n_i$ be the number of samples in cluster/class $i$, $\hat{n}_i$ be the number of samples clustered/classified by a method to cluster/class $i$, and $n_{i,j}$ be the number of samples in cluster/class $i$ clustered/classified by a method to cluster/class $j$. Then, the NMI measure for binary clustering is defined by
\begin{equation*}
  \mbox{NMI} = \frac{\sum_{i=1}^2 \sum_{j=1}^{2} n_{i,j} \log\left(n \cdot n_{i,j} / \left(n_i \cdot n'_j\right)\right)}{\sqrt{\left(\sum_{i=1}^2 n_i \log\left(n_i / n\right)\right)\left(\sum_{j=1}^{2} n'_j \log\left(n'_j / n\right)\right)}}.
\end{equation*}
Next, let TP (true positive) denote the number of samples which are correctly identified as class 1, FP (false positive) denote the number of samples which are incorrectly identified as class 1, TN (true negative) denote the number of samples which are correctly identified as class 2, FN (false negative) denote the number of samples which are incorrectly identified as class 2. Then, the RI measure is defined by
\begin{equation*}
  {\rm RI} = \frac{{\rm TP} + {\rm TN}}{{\rm TP} + {\rm TN} + {\rm FP} + {\rm FN}}.
\end{equation*}
As the obtained clustering and classification become more accurate, the values of NMI and RI become large.

\subsection{Artificial Data}
\begin{figure}[h]
  \begin{center}
    \includegraphics[scale=0.28]{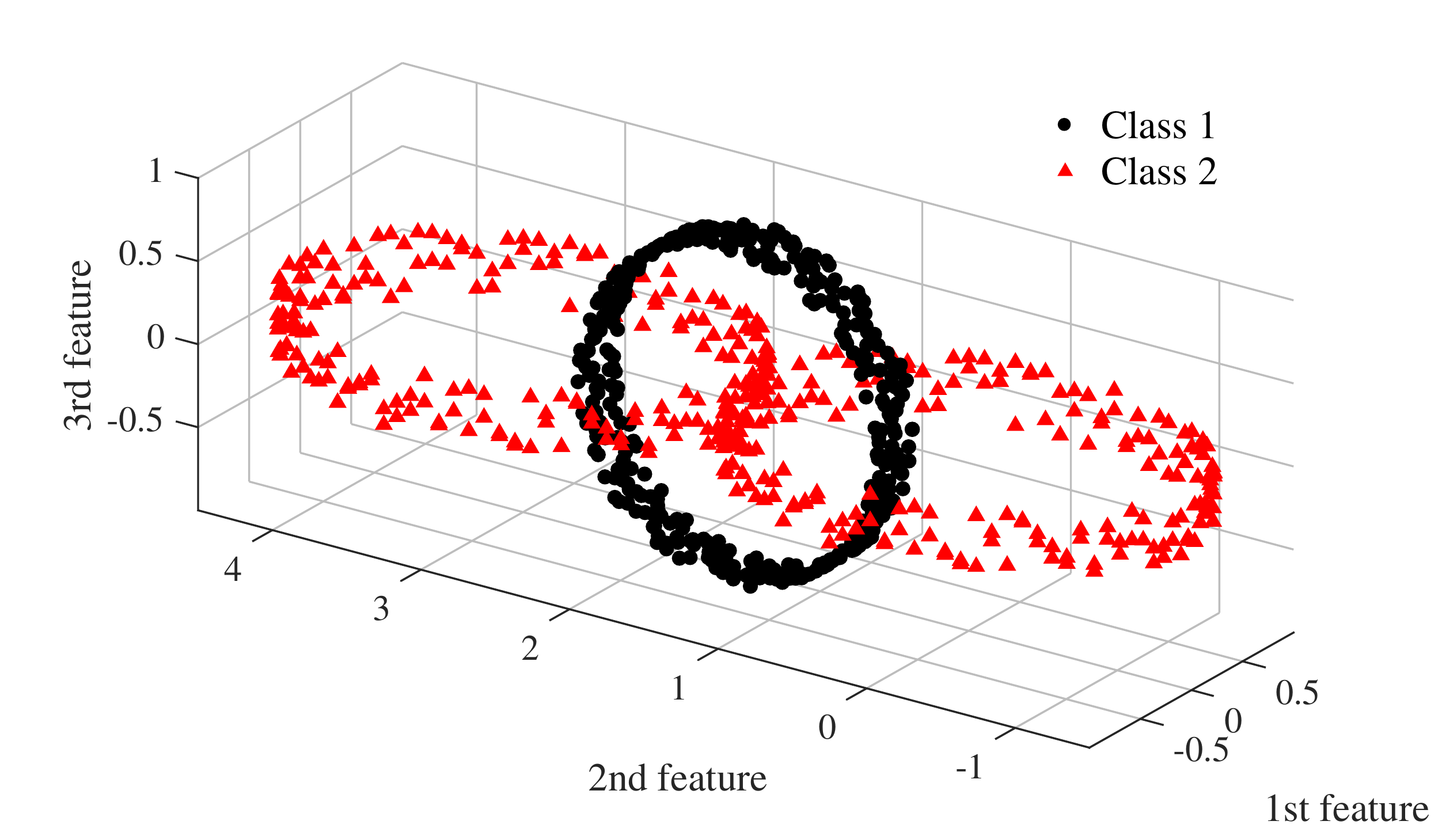} 
    \caption{Visualization in three dimensions of the artificial data}
    \label{fig:arti}
  \end{center}
  \begin{minipage}{0.47\hsize}
    \vspace{5pt}
  \begin{center}
    \includegraphics[scale = 0.47]{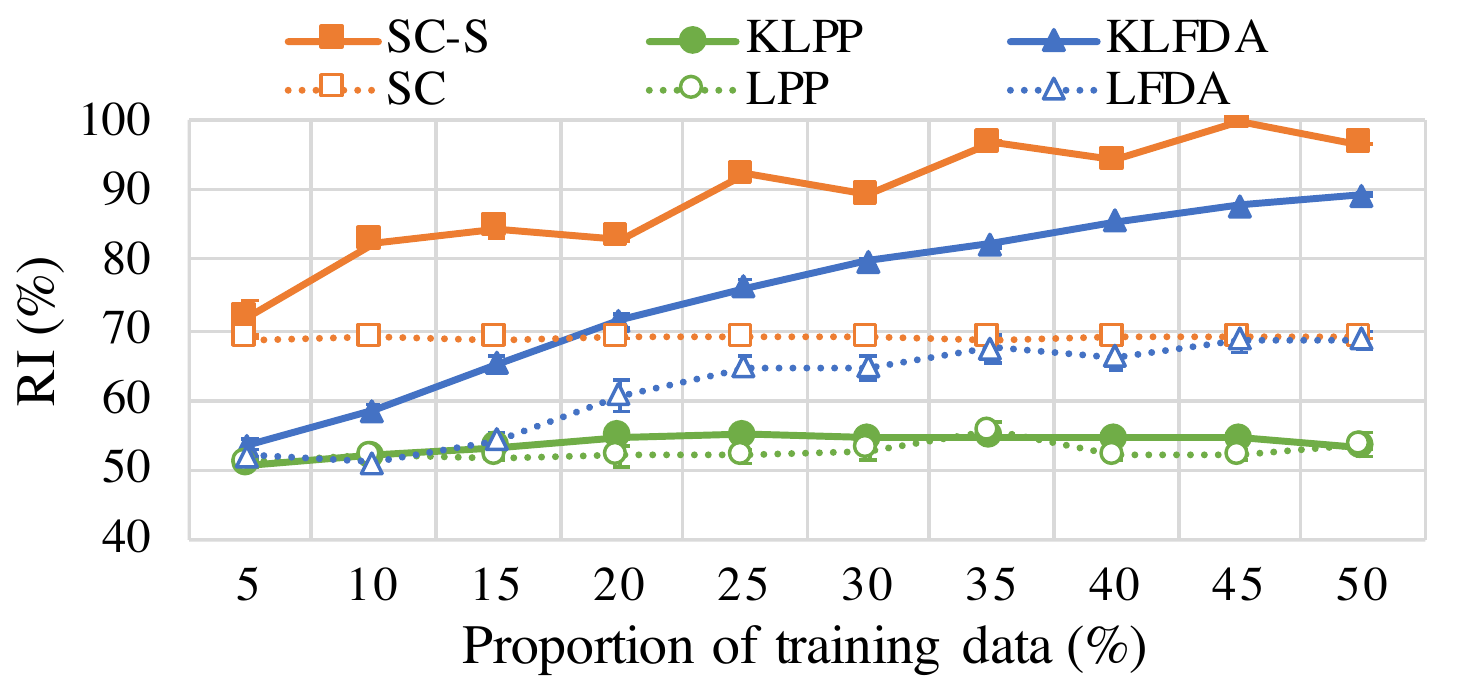} 
    \caption{{\small Proportion of training data vs.\ RI for clustering artificial data}}
    \label{fig:ACC}
  \end{center}
\end{minipage}
        \hspace{20pt}  
    \begin{minipage}{0.47\hsize}
      \begin{center}
        \includegraphics[scale = 0.47]{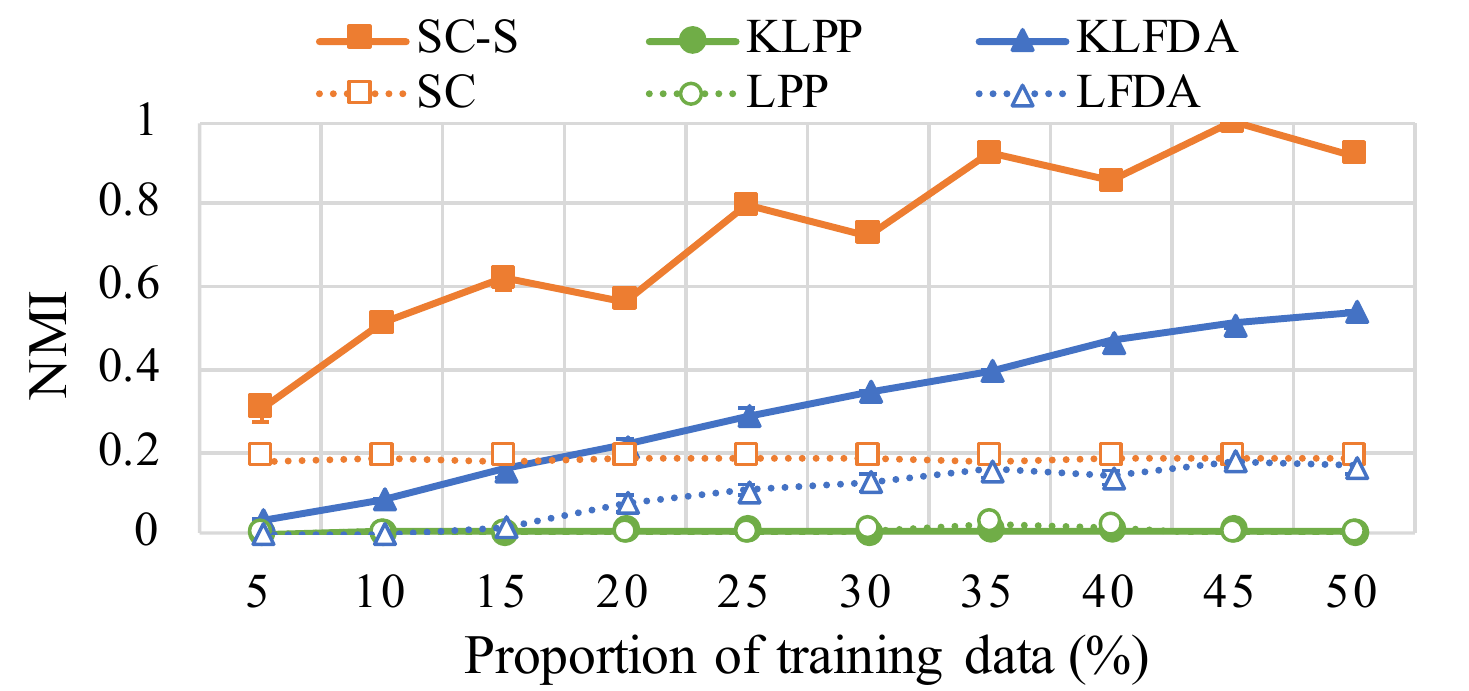} 
        \caption{{\small Proportion of training data vs.\ NMI for clustering artificial data}}
        \label{fig:NMI}
      \end{center}
    \end{minipage}
  \begin{minipage}{0.47\hsize}
      \vspace{15pt}
    \begin{center}
    \includegraphics[scale=0.46]{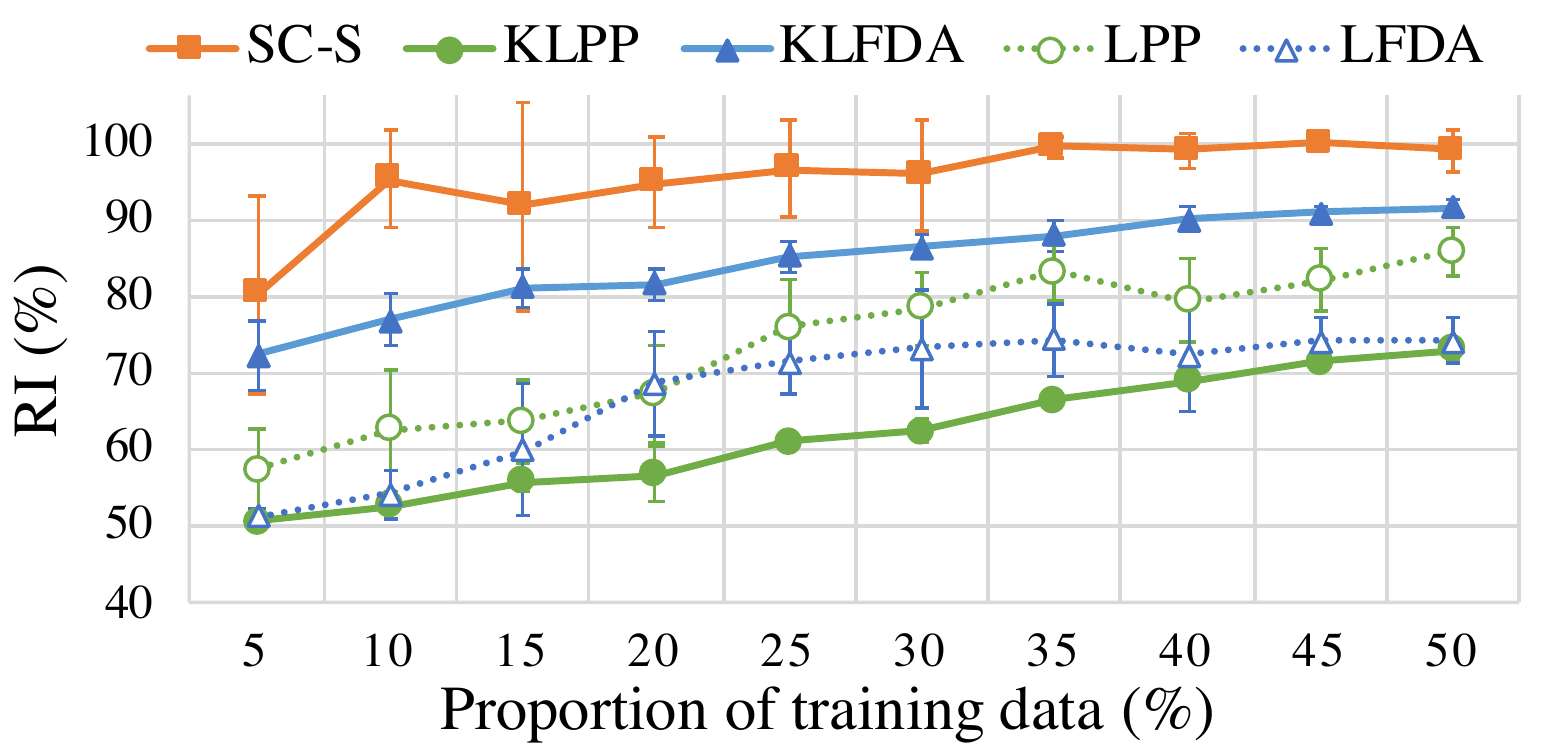}\\
    (a) One dimension
  \end{center}
\end{minipage}
\begin{minipage}{0.47\hsize}
  \begin{center}
    \includegraphics[scale=0.46]{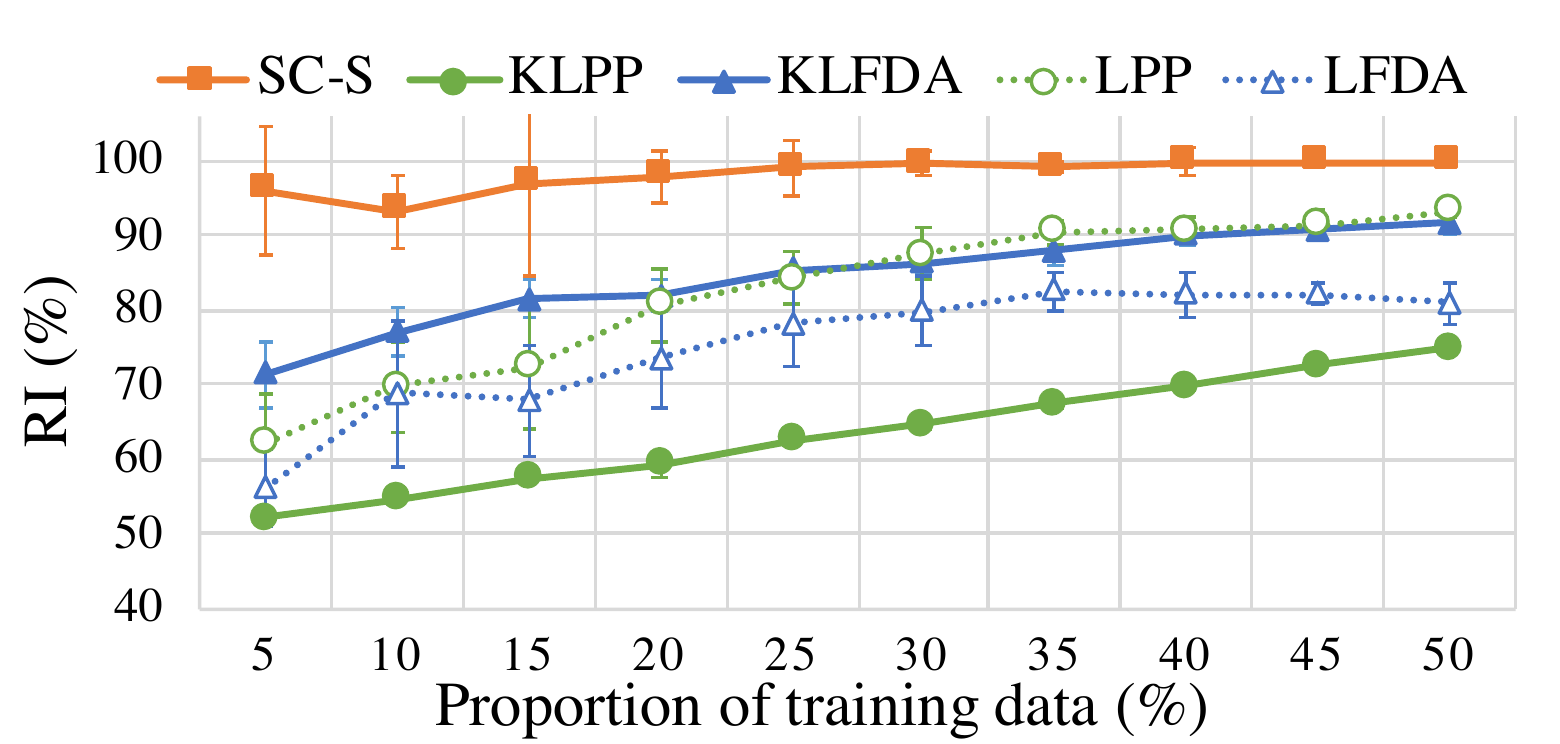}\\
    (b) Two dimensions
  \end{center}
\end{minipage}
\begin{minipage}{1\hsize}
  \begin{center}
    \includegraphics[scale=0.46]{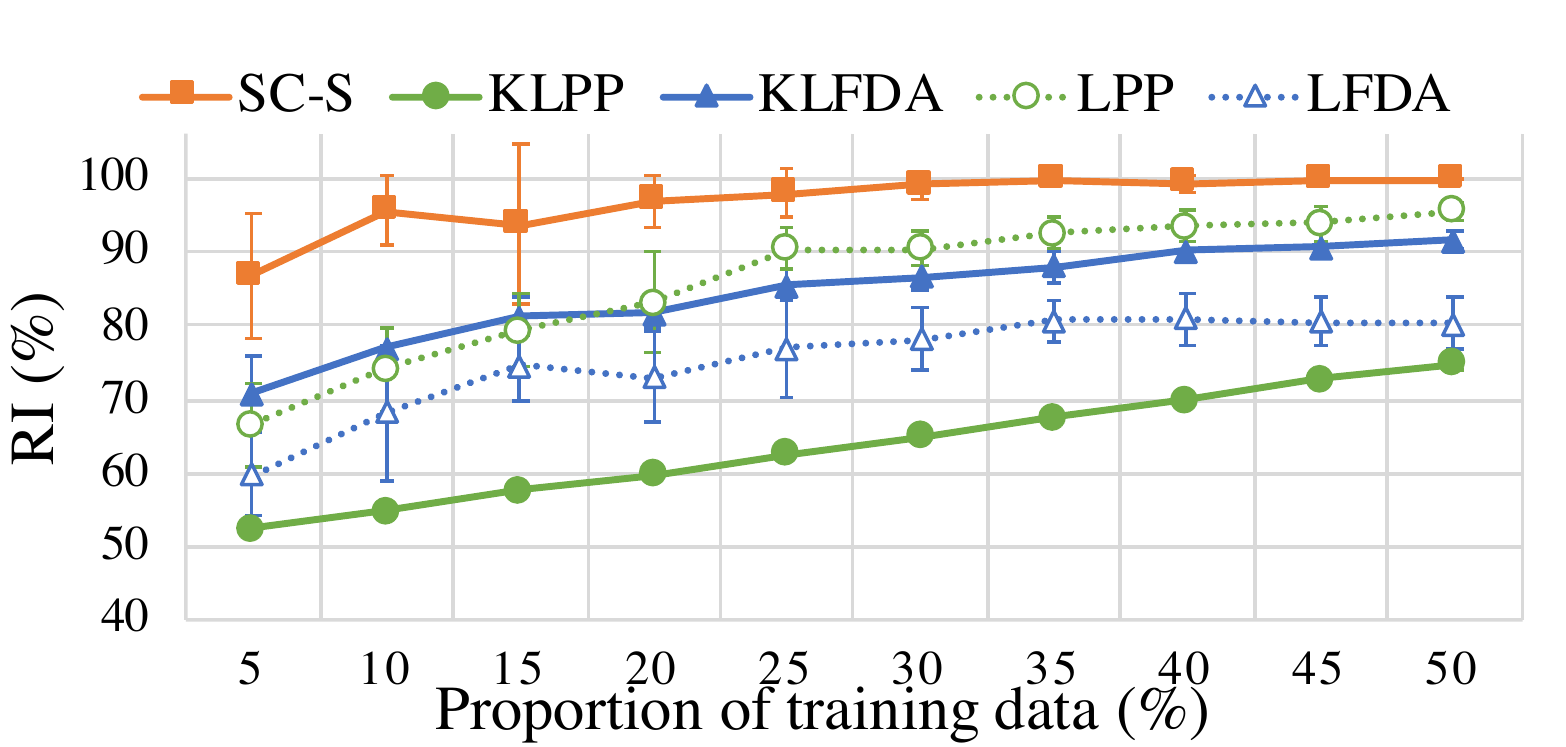}\\
    (c) Three dimensions
    \caption{Proportion of training data vs.\ RI for classification of artificia.l data for reduced dimensions $\ell = 1, 2,$ and $3$}
    \label{fig:KNN}
  \end{center}
  \end{minipage}
\end{figure}

In this subsection, we use toy problems of data sets consisting of 800 samples with 10 features. Figure~\ref{fig:arti} shows three of the 10 features, where the symbols $\circ$ and + denote the data samples in different classes, and the remaining seven features were given uniformly distributed random numbers in the interval $\left[0, 1 \right]$. The original data were normalized to have a mean of zero and variance one. We set the entries of the Fiedler vector $\bm{v}$ to $v_i = 1$ if sample $i$ is in a cluster, and $v_i = -0.2$ if sample $i$ is in another cluster (see (\ref{eq:j})). We repeated clustering the same reduced data samples 20 times using the $k$-means algorithm and the one-nearest neighborhood method, starting with different random initial guesses. Moreover, we tested on different proportions of training data from 5\% to 50\%, and repeated each test 10 times with different random choices of training data. Thus, we report the average performance by taking the arithmetic mean of RI and NMI.

Figures \ref{fig:ACC} and \ref{fig:NMI} show the means and standard deviations of RI and NMI, respectively, for the clustering problems. SC-S performed the best in terms of accuracy among the compared methods. The value of RI for SC-S, LFDA, and KLFDA tended to increase with the proportion of training data used. On the other hand, the accuracies of clustering using SC, LPP, and KLPP were similar for different proportions of training data.

Figure \ref{fig:KNN} shows the means and standard deviations of RI for the classification problems, where the reduced dimensions are $\ell = 1,2, \mbox{and}, 3$. SC-S performed the best in terms of accuracy for proportions of training data of 5--50\% with few exceptions among the compared methods. As the proportion of training data increased, the accuracy of the compared methods tended to increase.

\begin{figure}
\vspace{10pt}
  \begin{center}
    \includegraphics[scale=0.4]{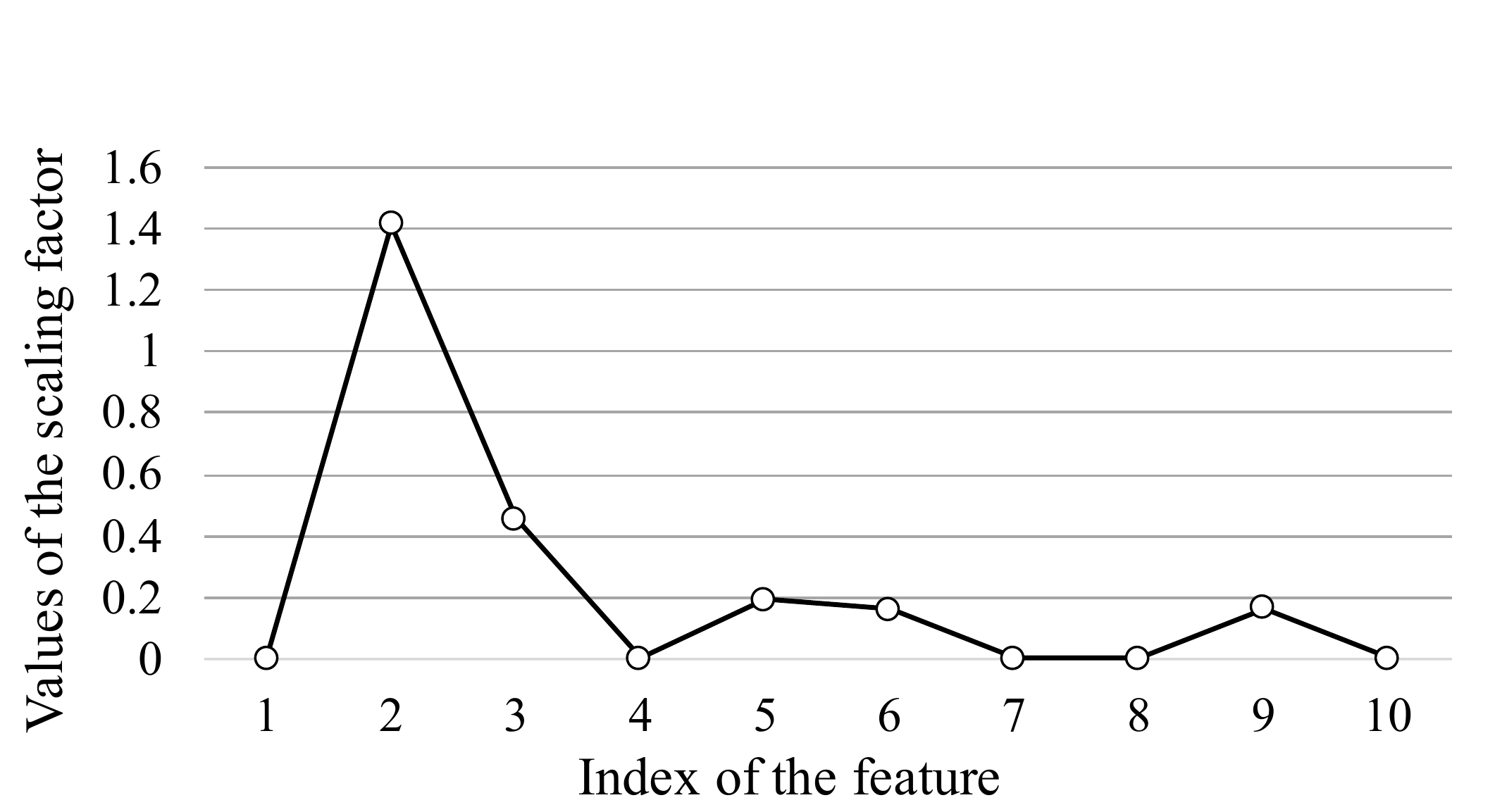}
    \caption{Values of the scaling factors}
    \label{fig:sf}
  \end{center}
\end{figure}

Figure \ref{fig:sf} shows the values of the scaling factors for each feature when using 50\% of the entire data for training. Since the fourth to tenth features given by random numbers were not involved in the classification, the values of the scaling factors of these seven features were small. Thus, the values of the scaling factors indicated which features were effective for the desired classification.

\begin{figure}
  \begin{minipage}{0.47\hsize}
  \begin{center}
    \includegraphics[scale = 0.4]{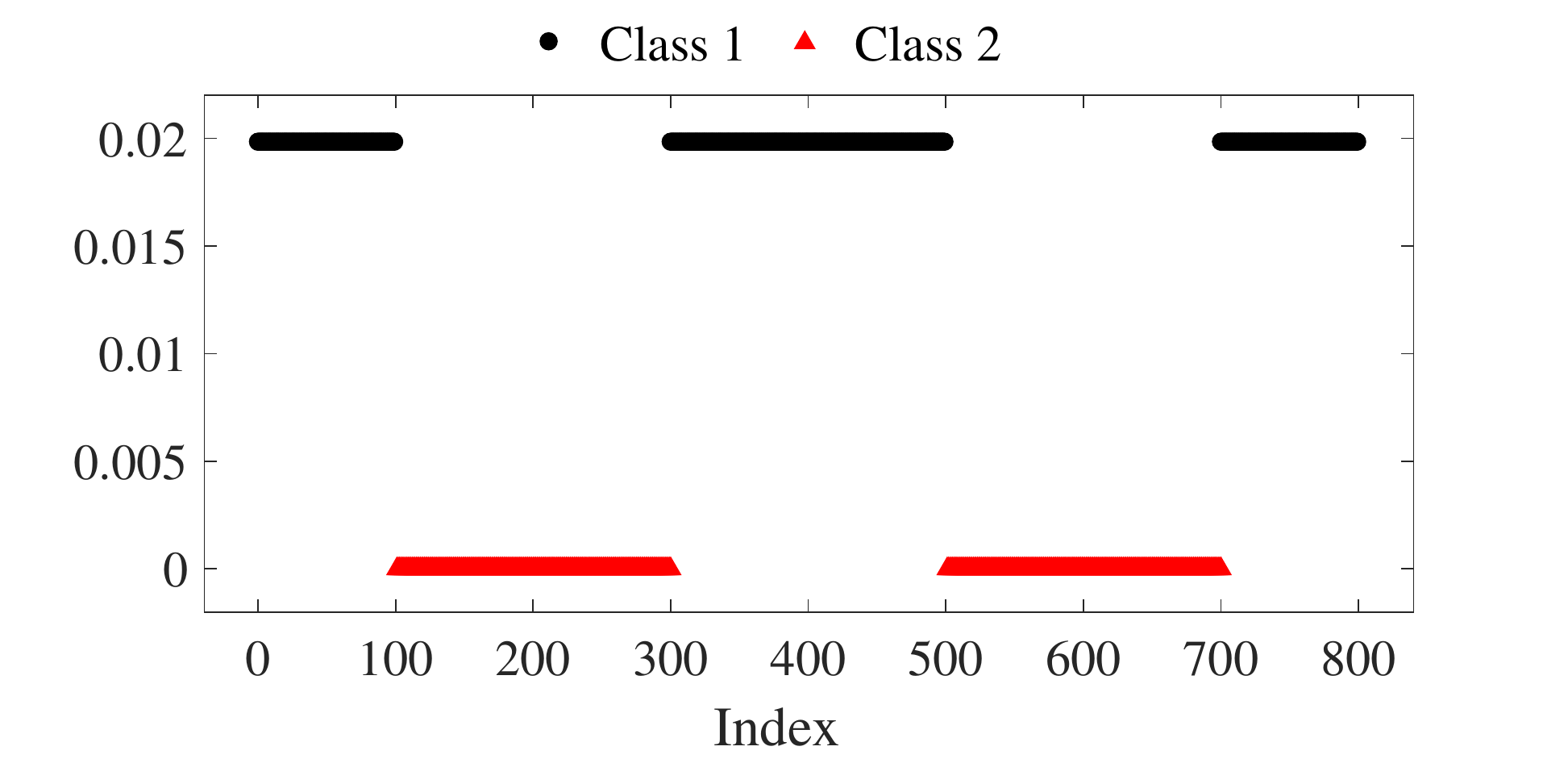}\\
    \vspace{2pt}
    (a) SC-S\\
  \end{center}
\end{minipage}
\hspace{20pt}
\begin{minipage}{0.47\hsize}
  \begin{center}
    \includegraphics[scale = 0.4]{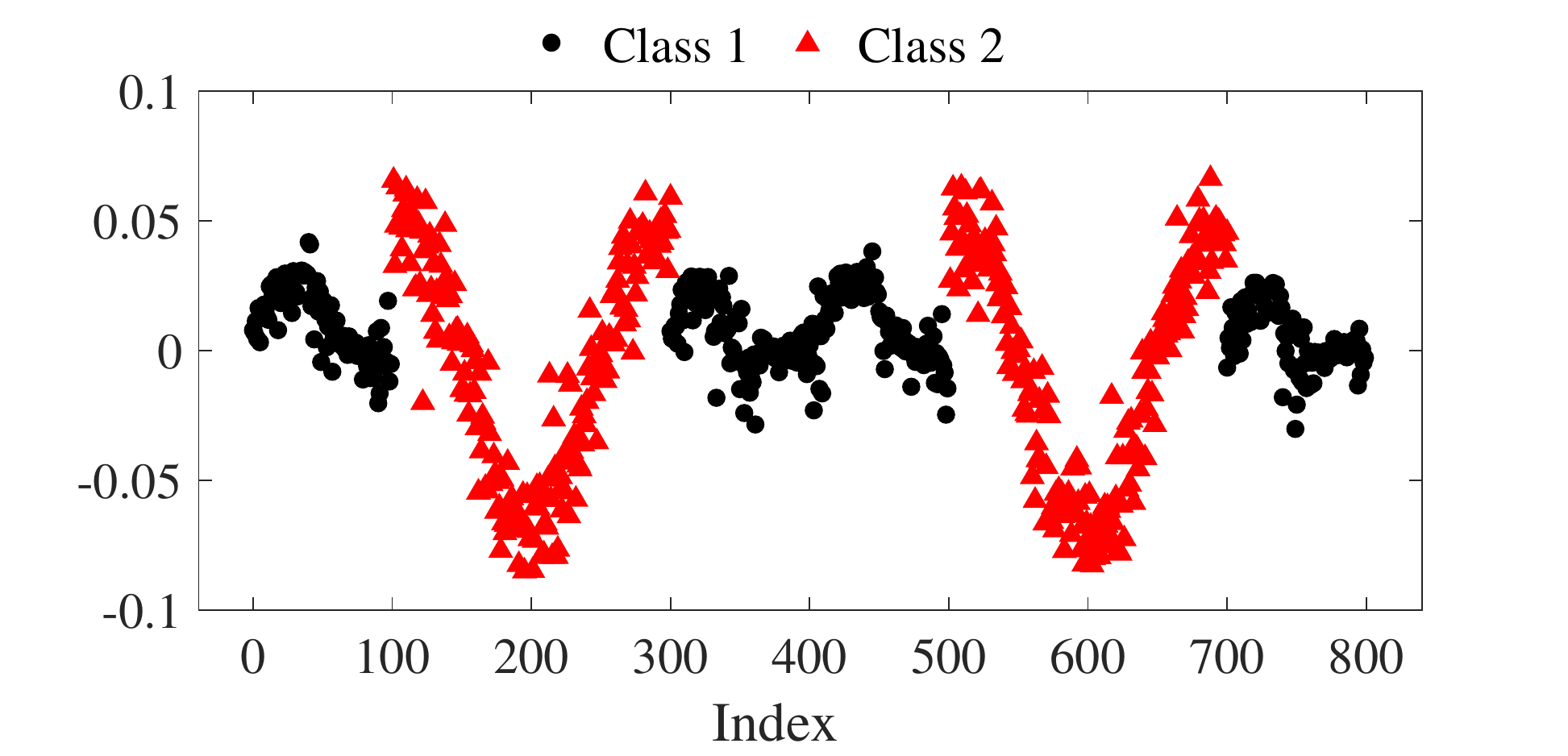}\\
    \vspace{2pt}
    (b) SC\\
  \end{center}
\end{minipage}
\begin{center}
    \vspace{5pt}
    \includegraphics[scale = 0.4]{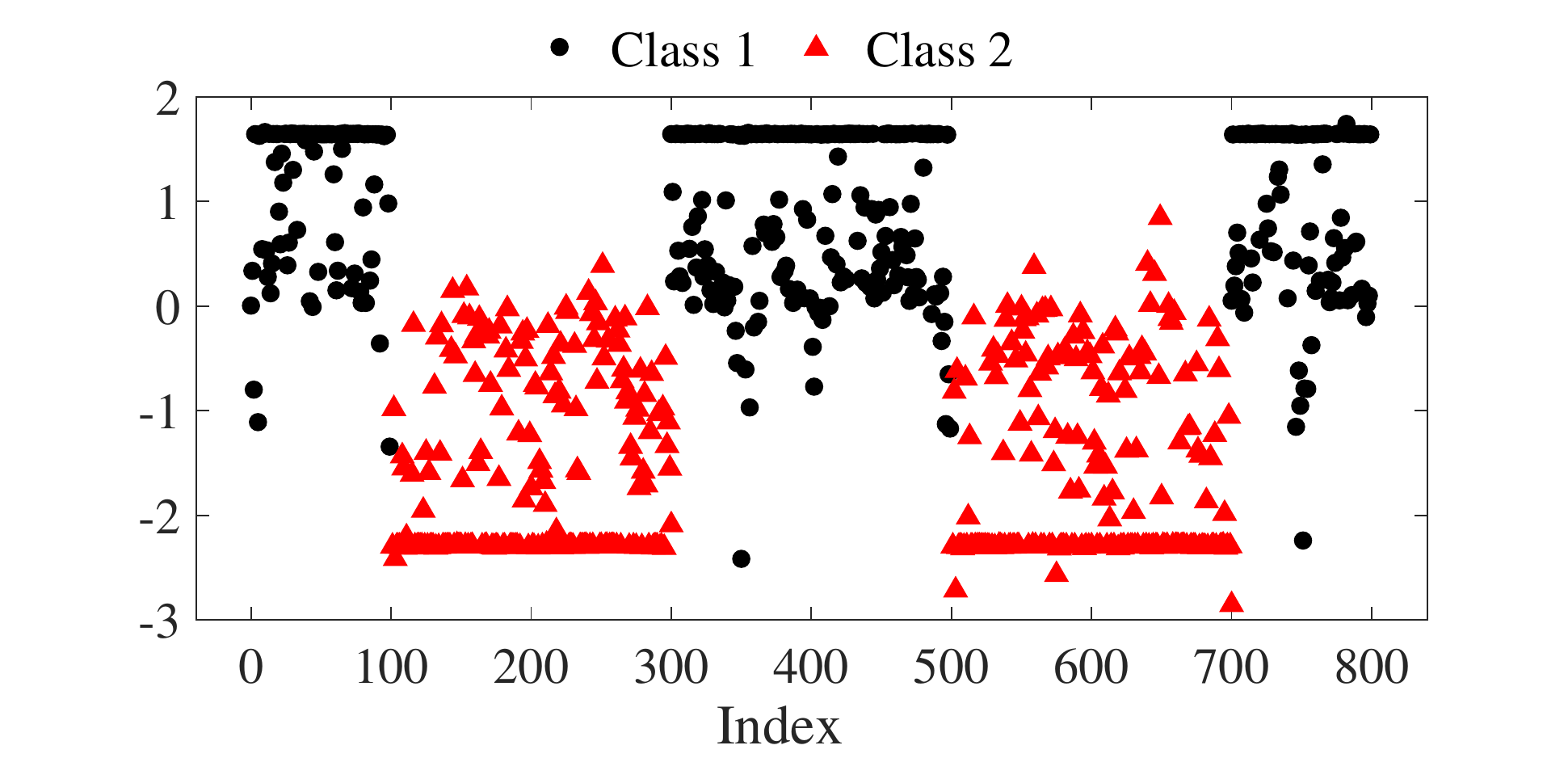}\\
    \vspace{2pt}
    (c) KLFDA
    \caption{Values of the reduced one-dimensional data samples}
    \label{fig:onespace}
  \end{center}
\end{figure}

Figure \ref{fig:onespace} shows the data samples reduced to a one-dimensional space for each method when using 50\% of all data for training. SC-S reduced all data to one dimension, and made it linearly separable to the true clusters. KLFDA failed to reduce the data samples to be linearly separable to the true clusters.

Table~\ref{tb:arti} gives the means and standard deviations of RI through the leave-one-out cross validation for the classification problems for reduced dimensions $\ell = 1,2,$ and $3$. Table~\ref{tb:arti} shows that SC-S was immune from an overfitting in the learning process.

\begin{table}[p]
  {\small
  \caption{RI (mean\% $\pm$ std) for classification of artificial data sets}
  \label{tb:arti}
  \begin{center}
    \begin{tabular}{p{32pt}|ccc}
      $\quad \ell$ & 1 & 2 & 3\\ \hline
      SC-S & *100.0$\pm$\phantom{0}0.0 & *100.0$\pm$\phantom{0}0.0  & *100.0$\pm$\phantom{0}0.0 \\
      LPP & \phantom{*0}77.0$\pm$40.1 & \phantom{*0}89.6$\pm$30.5 & \phantom{0*}99.9$\pm$\phantom{0}3.5\\
      KLPP & \phantom{*0}52.4$\pm$49.9 & \phantom{*0}52.4$\pm$50.0 & \phantom{*0}52.4$\pm$49.9\\
      LFDA & \phantom{*0}78.4$\pm$41.2 & \phantom{*0}83.9$\pm$36.8 & \phantom{*0}81.4$\pm$38.9\\
      KLFDA & \phantom{*0}86.8$\pm$33.9 & \phantom{*0}86.9$\pm$33.8 & \phantom{*0}87.0$\pm$33.6
    \end{tabular}
  \end{center}
  }
\end{table}

\newpage
\subsection{GEO Data Sets}
In this subsection, we use four data sets in the Gene Expression Omunibus\footnote{https://www.ncbi.nlm.nih.gov/} from real-world problems with more features than samples. Table~\ref{tb:geo} gives the specifications of the data sets, where IPF denotes Idiopathic Pulmonary Fibrosis. Each data set has two binary classes: diseased patients and healthy peoples. We chose 50\% of the entire data for training and repeated each test 10 times with different choices of the training data. The entries of the Fiedler vector $\bm{v}$ were set to $v_i = 1$ if sample $i$ is in a cluster, and $v_i = -1$ if sample $i$ is in another cluster (see (\ref{eq:j})).

Tables \ref{tb:accresult} and \ref{tb:nmiresult} give the means and standard deviations of RI and NMI, respectively, for each data set for the clustering problems. The symbol $\ast$ denotes the largest values of RI and NMI for each data set. SC-S performed the best in terms of accuracy for two data sets, and second best for two data sets among the compared methods. LFDA performed the best in terms of accuracy for one data set, second best for one data set but worst for one data set among the compared methods. These results showed that SC-S was more robust than other methods.

Table \ref{tb:error} gives the means and standard deviations of RI for the classification problems for the reduced dimensions $\ell = 1, 2,$ and $3$. SC-S was best in average in terms of accuracy for three data sets, except for Ovarian with $\ell = 1$. LFDA was best in terms of accuracy for Ovarian with $\ell = 1$, and for IPF with $\ell = 3$. KLFDA was best in terms of accuracy for IPF. SC-S became more accurate, as the number of eigenvectors $\ell$ used increased for Ovarian.

It is known that the cross-validation is unreliable in small sample classification \cite{cross}. Indeed, we observed significantly large values of the standard deviation for the compared methods for the data sets.

\begin{table}[h]
  \caption{Specifications of data sets}
  \label{tb:geo}
  \begin{center}
    \vspace{4pt}
    \begin{tabular}{l|c|c|c}
      \raisebox{4pt}{Data set} & \shortstack{\# of\\training samples} & \shortstack{\# of\\ entire samples} & \shortstack{\# of\\ features}\\[2pt] \hline
                               & \\ [-10pt]
      Ovarian  & 11 & 22 & 54675  \\
      Pancreatic & 12 & 24 & 54675  \\
      IPF & 12 & 23 & 54613  \\
      Colorectral & 11 & 22 & 54675  
    \end{tabular}
    \vspace{5pt}
  \end{center}

  \begin{minipage}{0.5\hsize}
    {\footnotesize
      \vspace{1pt}
  \caption{RI (mean\% $\pm$ std) for clustering GEO data sets}
  \label{tb:accresult}
  \begin{center}
    \vspace{4pt}
    \begin{tabular}{l|cccc}
      Data set & Ovarian & Pancreatic \\ \hline
      SC & \phantom{*}52.6 $\pm$ 0.05 & \phantom{*}54.2 $\pm$ 0.00 \\ 
      SC-S & *60.2 $\pm$ 1.11 & \phantom{*}56.0 $\pm$ 1.22 \\
      LPP & \phantom{*}57.0 $\pm$ 1.47 & *59.5 $\pm$ 1.20\\
      KLPP & \phantom{*}54.6 $\pm$ 0.00 & \phantom{*}54.2 $\pm$ 0.00\\
      LFDA & \phantom{*}57.9 $\pm$ 1.77 & \phantom{*}55.1 $\pm$ 1.57\\
      KLFDA & \phantom{*}56.4 $\pm$ 0.05 & \phantom{*}51.0 $\pm$ 0.08\\
    \end{tabular}
    \vspace{20pt}

    \begin{tabular}{l|cc}
      Data set & IPF & Colorectal\\ \hline
      SC & \phantom{*}69.6 $\pm$ 0.00 & \phantom{*}50.0 $\pm$ 0.00 \\ 
      SC-S & \phantom{*}76.1 $\pm$ 0.00 & *74.1 $\pm$ 1.10 \\
      LPP & \phantom{*}76.7 $\pm$ 2.12 & \phantom{*}57.8 $\pm$ 1.33\\
      KLPP & \phantom{*}71.3 $\pm$ 0.00 & \phantom{*}53.2 $\pm$ 0.00\\
      LFDA & \phantom{*}70.0 $\pm$ 4.34 & \phantom{*}54.3 $\pm$ 2.01\\
      KLFDA & *79.9 $\pm$ 0.12 & \phantom{*}63.4 $\pm$ 0.25\\
    \end{tabular}
  \end{center}
  }
\end{minipage}
\begin{minipage}{0.5\hsize}
  {\footnotesize
    \caption{NMI (mean $\pm$ std) for clustering GEO data sets}
    \label{tb:nmiresult}
    \begin{center}
      \vspace{4pt}
      \begin{tabular}{l|cccc}
        Data set & Ovarian & Pancreatic \\ \hline
        SC & \phantom{*}0.004 $\pm$ 0.000 & \phantom{*}0.043 $\pm$ 0.000 \\
        SC-S & *0.082 $\pm$ 0.019 & \phantom{*}0.042 $\pm$ 0.010 \\
        LPP & \phantom{*}0.044 $\pm$ 0.024 & *0.053 $\pm$ 0.022 \\
        KLPP & \phantom{*}0.006 $\pm$ 0.000 & \phantom{*}0.039 $\pm$ 0.000\\ 
        LFDA & \phantom{*}0.057 $\pm$ 0.021 & \phantom{*}0.040 $\pm$ 0.021\\
        KLFDA & \phantom{*}0.067 $\pm$ 0.001 & \phantom{*}0.005 $\pm$ 0.000\\
      \end{tabular}
      \vspace{20pt}

      \begin{tabular}{l|cc}
        Data set & IPF & Colorectal\\ \hline
        SC & \phantom{*}0.024 $\pm$ 0.000 & \phantom{*}0.041 $\pm$ 0.000\\ 
        SC-S & \phantom{*}0.093 $\pm$ 0.000 & *0.323 $\pm$ 0.016 \\
        LPP & \phantom{*}0.040 $\pm$ 0.014 & \phantom{*}0.048 $\pm$ 0.024\\
        KLPP & \phantom{*}0.040 $\pm$ 0.000 & \phantom{*}0.052 $\pm$ 0.000\\
        LFDA & \phantom{*}0.077 $\pm$ 0.041 & \phantom{*}0.030 $\pm$ 0.013\\
        KLFDA & *0.148 $\pm$ 0.005 & \phantom{*}0.109 $\pm$ 0.003\\
      \end{tabular}    
    \end{center}
  }
  \end{minipage}
\end{table}

\begin{table}[!ht]
\vspace{20pt}
  {\footnotesize
      \caption{RI (mean\% $\pm$ std) for classification of GEO data sets}
      \label{tb:error}
      \begin{minipage}{0.5\hsize}
    \begin{center}
      (a) Ovarian\\
      \vspace{5pt}
      \begin{tabular}{p{30pt}|ccc}
        \quad \ $\ell$ & 1 & 2 & 3\\ \hline
        SC-S & \phantom{*}75.0 $\pm$ 6.82 & *90.5 $\pm$ 6.88 & *90.0 $\pm$ 6.03\\
        LPP & \phantom{*}76.4 $\pm$ 4.90 & \phantom{*}75.5 $\pm$ 7.93 & \phantom{*}75.9 $\pm$ 8.87\\
        KLPP & \phantom{*}55.5 $\pm$ 4.45 & \phantom{*}56.4 $\pm$ 6.17 & \phantom{*}57.7 $\pm$ 5.40\\
        LFDA & *77.3 $\pm$ 7.87 & \phantom{*}70.9 $\pm$ 4.64 & \phantom{*}69.6 $\pm$ 5.40\\
        KLFDA & \phantom{*}55.9 $\pm$ 6.12 & \phantom{*}55.0 $\pm$ 3.78 & \phantom{*}56.4 $\pm$ 4.17\\
      \end{tabular}
    \end{center}
  \end{minipage}
  \begin{minipage}{0.5\hsize}
    \begin{center}
      (b) Pancreatic\\
      \vspace{5pt}
      \begin{tabular}{p{30pt}|ccc}
        \quad \ $\ell$ & 1 & 2 & 3\\ \hline
        SC-S & *77.5 $\pm$ 8.58 & *77.9 $\pm$ 7.69 & *73.3 $\pm$ 7.50\\
        LPP & \phantom{*}74.6 $\pm$ 8.00 & \phantom{*}73.8 $\pm$ 5.61 & \phantom{*}72.1 $\pm$ 7.23\\
        KLPP & \phantom{*}55.4 $\pm$ 4.19&\phantom{*}56.7 $\pm$ 3.33 & \phantom{*}54.6 $\pm$ 3.46\\
        LFDA &\phantom{*}50.0 $\pm$ 0.00&\phantom{*}50.0 $\pm$ 0.00 & \phantom{*}50.0 $\pm$ 0.00\\
        KLFDA & \phantom{*}54.2 $\pm$ 2.64 & \phantom{*}54.2 $\pm$ 2.64 & \phantom{*}54.2 $\pm$ 2.64\\
      \end{tabular}
    \end{center}
  \end{minipage}
  \begin{minipage}{0.5\hsize}
    \begin{center}
      \vspace{20pt}
      (c) IPF\\ 
      \vspace{5pt}
      \begin{tabular}{p{30pt}|ccc}
        \quad \ $\ell$ & 1 & 2 & 3\\ \hline
        SC-S & *90.4 $\pm$ 7.48 & *96.5 $\pm$ 3.79 & *97.4 $\pm$ 4.43\\
        LPP & \phantom{*}74.4 $\pm$ 9.01 & \phantom{*}76.5 $\pm$ 8.06 & \phantom{*}76.5 $\pm$ 8.30\\
        KLPP & \phantom{*}68.7 $\pm$ 3.25 & \phantom{*}65.2 $\pm$ 4.76 & \phantom{*}67.8 $\pm$ 9.76\\
        LFDA &\phantom{*}75.2 $\pm$ 8.26 & \phantom{*}84.8 $\pm$ 6.52 & \phantom{*}87.0 $\pm$ 0.00\\
        KLFDA & \phantom{*}87.0$\pm$ 0.00 & \phantom{*}87.0 $\pm$ 0.00 & \phantom{*}87.0 $\pm$ 0.00\\
      \end{tabular}
    \end{center}
  \end{minipage}
  \begin{minipage}{0.5\hsize}
    \begin{center}
      \vspace{20pt}
      (d) Colorectal\\
      \vspace{5pt}      
      \begin{tabular}{l|ccc}
        \quad \ $\ell$ & 1 & 2 & 3\\ \hline
        SC-S & *82.3 $\pm$ 8.73 & *85.5 $\pm$ 6.98 & *82.7 $\pm$ 6.98\\
        LPP & \phantom{*}72.3 $\pm$ 9.63 & \phantom{*}75.0 $\pm$ 6.18 & \phantom{*}80.0 $\pm$ 4.64\\
        KLPP & \phantom{*}57.3 $\pm$ 7.39 & \phantom{*}60.9 $\pm$ 6.80 & \phantom{*}64.6 $\pm$ 7.27\\
        LFDA & \phantom{*}70.5 $\pm$ 3.05 & \phantom{*}77.3 $\pm$ 4.07 & \phantom{*}79.1 $\pm$ 5.82\\
        KLFDA & \phantom{*}77.3 $\pm$ 0.00 & \phantom{*}77.3 $\pm$ 0.00 & \phantom{*}77.3 $\pm$ 0.00\\
      \end{tabular}
    \end{center}
    \end{minipage}
    }
\end{table}

\section{Conclusions}
We considered the dimensionality reduction of high-dimensional data. To deal with irregularity or uncertainty in features, we proposed supervised dimensionality reduction methods that exploit knowledge on the labels of partial samples. With the supervision, we modify the features' variances and means. Moreover, the feature scaling can reduce those features that prevent us from obtaining the desired clusters. To obtain the factors used to scale the features, we formulated an eigenproblem of a linear matrix pencil whose eigenvector has the feature scaling factors, and described the procedures for the proposed methods. Numerical experiments showed that the feature scaling is effective when combined with spectral clustering and classification methods. For toy problems with more samples than features, the feature scaling improved the accuracy of the unsupervised spectral clustering, and the proposed methods outperformed several existing methods. For real-world problems with more features than samples from gene expression profiles, the proposed method was more robust in terms of clustering than well-established methods, and outperformed existing methods in some cases.

\section*{Acknowledgements}
We would like to thank Doctor Akira Imakura for his valuable comments. The second author was supported in part by JSPS KAKENHI Grant Number 15K04768 and the Hattori Hokokai Foundation. The third author was supported by JST/CREST, Japan.

\bibliographystyle{ieeetr}
\bibliography{arXiv}

\end{document}